\documentclass[runningheads]{llncs}

\usepackage{eccv}

\usepackage{eccvabbrv}

\usepackage{graphicx}
\usepackage{booktabs}
\usepackage{authblk}

\usepackage[accsupp]{axessibility}  

\usepackage[pagebackref,breaklinks,colorlinks,citecolor=eccvblue]{hyperref}

\usepackage{orcidlink}

\usepackage{pifont}
\usepackage{tikz}
\usepackage{bbding}
\usepackage[flushleft]{threeparttable}
\usepackage{xr}
\makeatletter
\newcommand*{\addFileDependency}[1]{
  \typeout{(#1)}
  \@addtofilelist{#1}
  \IfFileExists{#1}{}{\typeout{No file #1.}}
}
\makeatother

\newcommand*{\myexternaldocument}[1]{
    \externaldocument{#1}
    \addFileDependency{#1.tex}
}
\myexternaldocument{supplementary}

\begin{document}

\title{Every Pixel Has its Moments: Ultra-High-Resolution Unpaired Image-to-Image Translation via Dense Normalization} 

\titlerunning{Dense Normalization}


\author{Ming-Yang Ho\inst{1}\and
Che-Ming Wu\inst{2} \and
Min-Sheng Wu\inst{3} \and Yufeng Jane Tseng \inst{1}}

\authorrunning{Ho. et al.}

\institute{
National Taiwan University \and
Amazon Web Services \and
aetherAI\\
\email{kaminyou@cmdm.csie.ntu.edu.tw, unowu@amazon.com, vincentwu@aetherai.com, yjtseng@csie.ntu.edu.tw}
}



\maketitle

\begin{abstract}
Recent advancements in ultra-high-resolution unpaired image-to-image translation have aimed to mitigate the constraints imposed by limited GPU memory through patch-wise inference. Nonetheless, existing methods often compromise between the reduction of noticeable tiling artifacts and the preservation of color and hue contrast, attributed to the reliance on global image- or patch-level statistics in the instance normalization layers. In this study, we introduce a Dense Normalization (DN) layer designed to estimate pixel-level statistical moments. This approach effectively diminishes tiling artifacts while concurrently preserving local color and hue contrasts. To address the computational demands of pixel-level estimation, we further propose an efficient interpolation algorithm. Moreover, we invent a parallelism strategy that enables the DN layer to operate in a single pass. Through extensive experiments, we demonstrate that our method surpasses all existing approaches in performance. Notably, our DN layer is hyperparameter-free and can be seamlessly integrated into most unpaired image-to-image translation frameworks without necessitating retraining. Overall, our work paves the way for future exploration in handling images of arbitrary resolutions within the realm of unpaired image-to-image translation. Code is available at: \url{https://github.com/Kaminyou/Dense-Normalization}.

  \keywords{Unpaired image-to-image translation \and Ultra-high-resolution image \and Parallelism}
\end{abstract}

\section{Introduction}
\label{sec:intro}

Unpaired image-to-image (I2I) translation is a conventional computer vision task that aims to translate an image from one domain to another without using paired images~\cite{hoyez2022unsupervised, kaji2019overview, pang2021image}. However, most frameworks are incapable of handling ultra-high-resolution (UHR) images due to GPU memory limitations. For example, a popular CUT~\cite{park2020contrastive} framework requires 14 GB of GPU VRAM for inference and 160 GB for training when processing an image with a resolution of 2,048$\times$2,048, exceeding the capacity of a single 32GB NVIDIA V-100 GPU. This presents a significant challenge for researchers and practitioners working on unpaired I2I translation tasks involving UHR images.

Nevertheless, the ubiquity of UHR images in our daily life is undeniable, with mobile phones capturing 4K resolution photos and movies exceeding 8K resolution~\cite{funatsu20156, kitamura2011beyond}. Without an effective methodology, performing common image translation tasks like style transfer~\cite{jing2019neural} and colorization~\cite{vzeger2021grayscale} on these images would be significantly hindered.

Another critical application of UHR unpaired I2I translation is stain transformation in digital pathology~\cite{de2021deep, yang2022virtual, zhang2022mvfstain}. Standard staining methods, such as hematoxylin and eosin (H\&E), are commonly used due to their cost-effectiveness. However, for more detailed cancer diagnostics, the use of expensive immunohistochemical (IHC) stains becomes essential~\cite{birkman2018gastric, inamura2018update}. Given that pathological images frequently have resolutions exceeding 10,000$\times$10,000 pixels, an effective algorithm for stain transformation that reduces the cost of pathological staining is urgently needed.


A few strategies have been leveraged to perform unpaired I2I translation on UHR images. Simplifying the model architecture or increasing the output image size enable translation on 2K images~\cite{song2022multi,liang2021high}, but they still have a high GPU memory usage with space complexity of $\mathcal{O}(N^2)$ for an image with a resolution of $N\times N$. Alternatively, patch-wise training and inference can decrease space complexity to $\mathcal{O}(1)$, but struggle to produce seamless results due to the tiling artifacts that appear when stitching the patches into an UHR image. Although convolutional operators in most I2I frameworks should guarantee that the final output can be seamlessly assembled from the patches, normalization operators applied per patch disrupt this property. Since the statistical moments calculated in Instance Normalization (IN) layers can affect color fidelity~\cite{huang2017arbitrary}, their discrepancies between neighboring patches lead to \textit{gap-type tiling artifacts}, evidenced by patch-wise IN~\cite{ulyanov2016instance} (refer to Fig.~\ref{fig:real2paint_large} (b) and Fig.~\ref{fig:concept} (a)).


{
\setlength\abovecaptionskip{-1.0ex}
\setlength\belowcaptionskip{-3.5ex}
\begin{figure}[t]
\begin{center}
\includegraphics[width=0.8\linewidth]{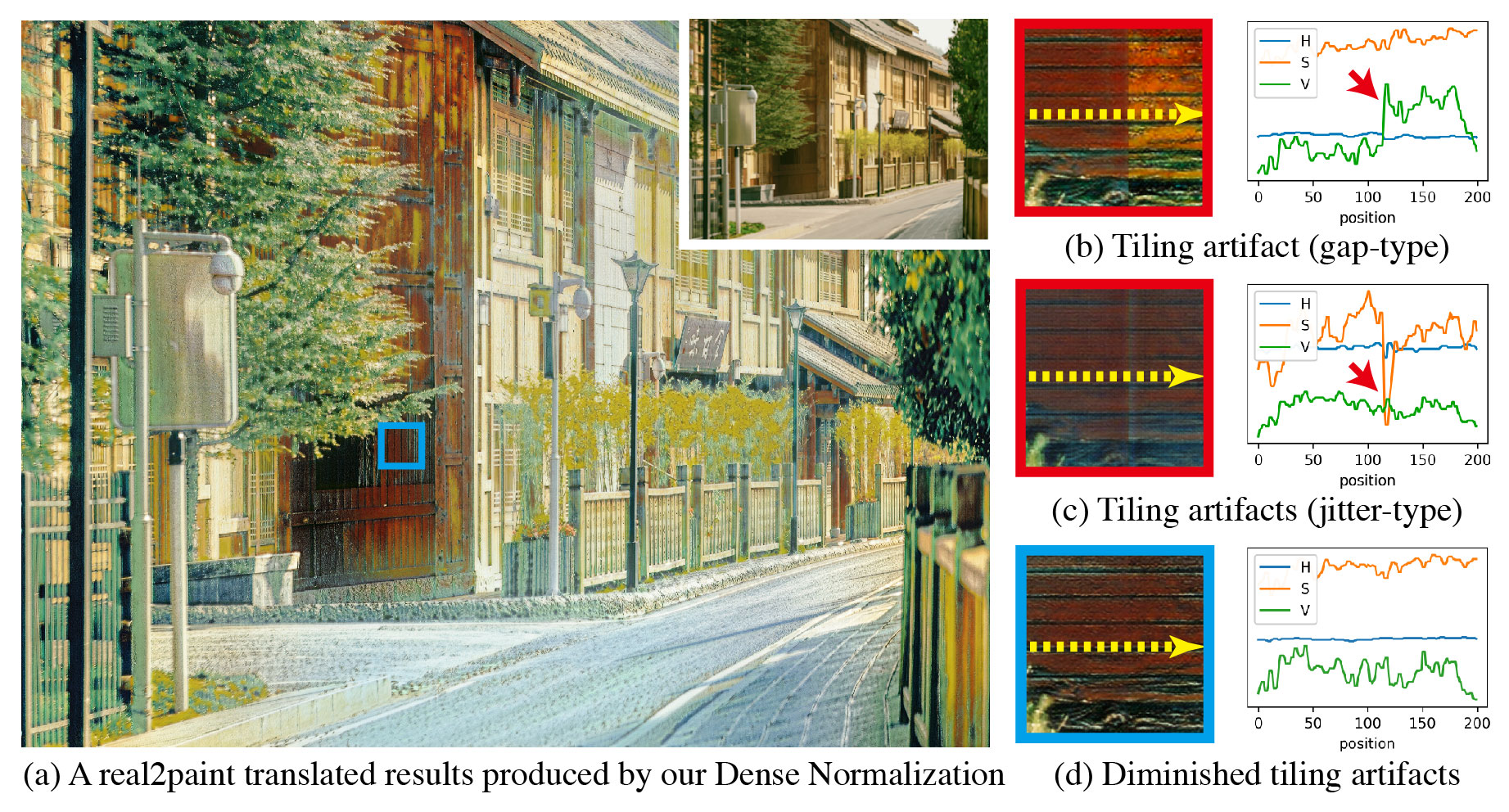}
\end{center}
   \caption{\textbf{Comparison of translations.} (a) Showcases a real2paint translated ultra-high-resolution image (3,024$\times$4,032 pixels) produced by our Dense Normalization (DN) from the image displayed in the top right corner, with comparisons highlighted within the blue-boxed region. (b) Illustrates the occurrence of gap-type tiling artifacts in patch-wise IN~\cite{ulyanov2016instance} or KIN~\cite{ho2022ultra}; (c) Demonstrates jitter-type tiling artifacts resulting from TIN~\cite{chen2022towards}; (d) Presents DN's effectiveness in diminishing tiling artifacts.}
\label{fig:real2paint_large}
\end{figure}
}

To mitigate this issue, Thumbnail Instance Normalization (TIN)\cite{chen2022towards} applies global image-level statistics, at the expense of losing local hue and contrast, resulting in over/under coloring. Furthermore, significant perturbations in these statistical moments unfortunately create \textit{jitter-type tiling artifacts}, which manifest as color jitters at the edges of patches (see Fig.~\ref{fig:real2paint_large} (c) and Fig.~\ref{fig:concept} (b)). On the other hand, Kernelized Instance Normalization (KIN)~\cite{ho2022ultra} alleviates gap-type tiling artifacts by more closely aligning adjacent patch-level statistics, but it necessitates selecting a kernel size to make a trade-off between blurring artifacts and preserving local color contrast (see Fig. S1 in the supplementary material). Additionally, KIN's method requires a two-stage pipeline (caching and inference stages) due to the initial need for statistics calculation and subsequent performance of convolution operations on them (see Fig.~\ref{fig:concept} (c)). This raises a question: \textit{Can pixel-level statistical moment estimation address all these issues, and can it be accomplished in a single pass?}

{
\setlength\abovecaptionskip{-0.5ex}
\setlength\belowcaptionskip{-4.0ex}
\begin{figure*}[t]
\begin{center}
\includegraphics[width=0.85\linewidth]{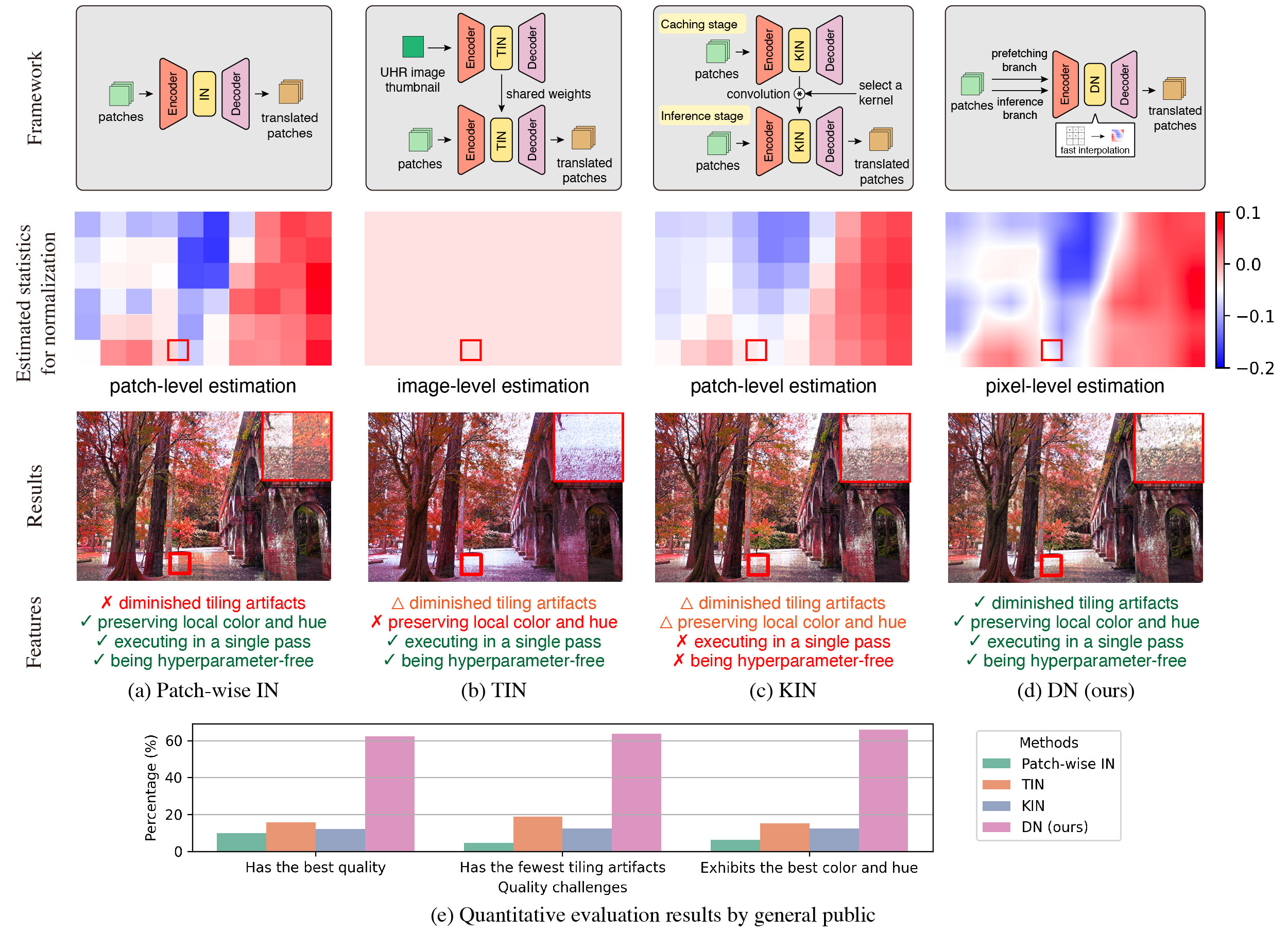}
\end{center}
   \caption{\textbf{Comparison of various normalization strategies.} This figure illustrates the framework and the impact of different normalization methods on an UHR image (3,024$\times$4,032 pixels) for the summer2autumn task: (a) Patch-wise IN~\cite{ulyanov2016instance} uses patch-level statistics and leads to statistical differences between patches, resulting in noticeable gap-type tiling artifacts. (b) TIN~\cite{chen2022towards} eliminates statistical differences with global image-level statistics (from the thumbnail) but compromises color and hue details, also inducing jitter-type tiling artifacts. (c) KIN~\cite{ho2022ultra} utilizes a two-stage pipeline to mitigate statistical differences by applying convolutional operations on patch-level statistics, albeit at the expense of local detail. (d) DN estimates pixel-level statistical moments in a single pass, effectively preserving local color and hue while diminishing tiling artifacts. (e) DN outperforms all methods in every aspect of human evaluation. In the row of features, \textcolor{teal}{\checkmark} indicates ``achieved''; \textcolor{orange}{$\bigtriangleup$} indicates ``partially achieved''; \textcolor{red}{\ding{55}} indicates ``not achieved''. Red close-up boxes highlight the outcomes influenced by different statistical moments used for normalization.}
\label{fig:concept}
\end{figure*}
}

To answer the above question, we propose the Dense Normalization (DN) layer, which is capable of estimating statistical moments for every pixel.  It possesses four expected properties: diminishing tiling artifacts ($\mathcal{P}_1$), preserving local hue and color contrast ($\mathcal{P}_2$), executing in a single pass ($\mathcal{P}_3$), and being hyperparameter-free ($\mathcal{P}_4$), as illustrated in Fig.\ref{fig:real2paint_large}, Fig.\ref{fig:concept}(d), and Fig. S2 in the supplementary material.

While pixel-level statistics estimation can be achieved by performing bilinear interpolation on patch-level statistics, its naïve implementation is time-consuming due to the high computational demands. Hence, we developed a \textit{fast interpolation} algorithm to enhance calculation efficiency and practicality (see the comparison in Table~\ref{tab:ablation_algorithm}). Furthermore, to perform pixel-level statistics estimation, patch-wise statistics must first be calculated and cached. Fast interpolation is then performed on these statistics, a process that would typically necessitate a two-stage pipeline similar to KIN. Nevertheless, we have devised a \textit{prefetching} strategy that cleverly hides the caching process within the inference process, leveraging GPU parallelism to enable DN to operate in a single pass.



We evaluated our DN on four publicly available datasets, including natural and pathological images; quantitative evaluations confirmed that DN outperforms previous methods and its applicability in the healthcare field is demonstrated. In summary, our research has achieved the following:
\begin{itemize}
    \item To the best of our knowledge, this is the first study to estimate pixel-level statistical moments for normalization in UHR unpaired I2I translation, effectively diminishing tiling artifacts ($\mathcal{P}_1$) while simultaneously preserving local hue and color ($\mathcal{P}_2$), thereby achieving state-of-the-art performance.
    \item We introduce a fast interpolation algorithm for efficient pixel-wise statistics estimation, along with a prefetching parallelism algorithm that enables DN to operate in a single pass ($\mathcal{P}_3$), significantly decreasing runtime in comparison to naïve implementations.
    \item Our hyperparameter-free DN layer ($\mathcal{P}_4$) can be seamlessly integrated into any existing framework utilizing IN layers during inference, without necessitating model retraining.
\end{itemize}

\section{Related works}
\noindent\textbf{Unpaired image-to-image translation.} Several frameworks have been developed for unpaired image-to-image translation, aiming to discover the mapping between diverse image domains. CycleGAN~\cite{CycleGAN2017}, DiscoGAN~\cite{kim2017learning}, and DualGAN~\cite{yi2017dualgan} utilize cycle-consistency loss to enforce the mapping. However, the pixel-level cycle-consistency constraint can lead to deformation and hinder the generation of large objects and fine textures when there are significant domain differences. Recently, strategies have been proposed to enhance performance beyond cycle-consistency. DistanceGAN~\cite{Benaim2017OneSidedUD} maintains pairwise distances between different parts of the same sample in each domain, while ACL-GAN~\cite{zhao2020unpaired} utilizes adversarial loss to address cyclic loss. CUT~\cite{park2020contrastive} maximizes patch-wise similarity between domains using contrastive learning, and LSeSim~\cite{zheng2021spatiallycorrelative} learns spatial correlation to preserve structural similarity. Additionally, patch-wise semantic relationship regularization~\cite{jung2022exploring} is used to enhance correspondence between input and output images, while an energy function~\cite{zhao2022egsde} is employed to retain domain-independent features and discard domain-specific ones. Despite these advancements, these frameworks are limited to processing small images. Our DN is a plugin designed to enable the processing of UHR images by simply replacing the IN layer in these frameworks.

{
\setlength\abovecaptionskip{-0.5ex}
\setlength\belowcaptionskip{-3.5ex}
\begin{figure*}[t]
\begin{center}
\includegraphics[width=0.90\linewidth]{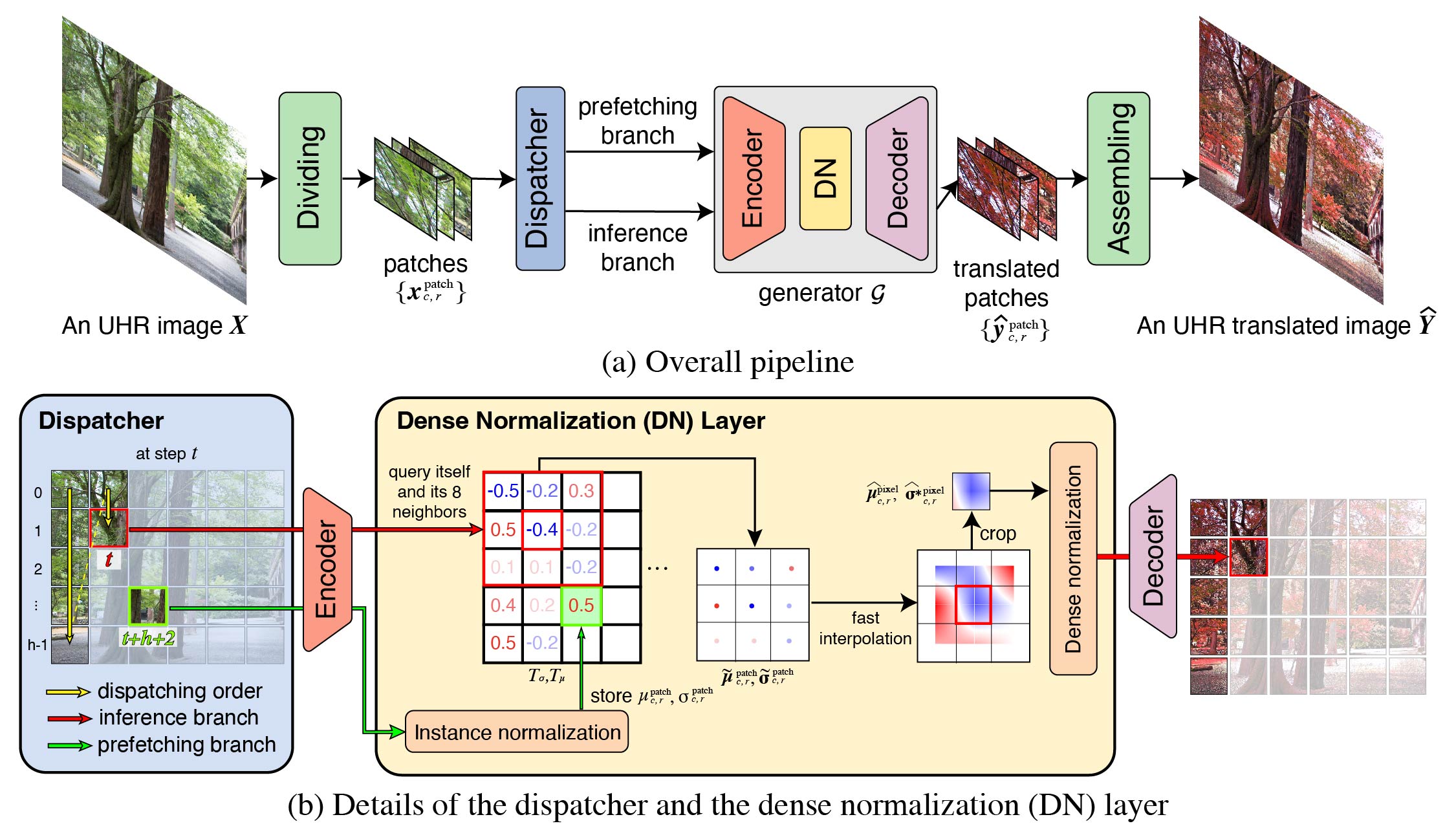}
\end{center}
   \caption{\textbf{Framework of the Proposed Method.} (a) Provides an overall view of our framework's pipeline. (b) Shows the details of the dispatcher and the Dense Normalization (DN) layer. A UHR image $\boldsymbol{X}$ is initially divided into patches $\boldsymbol{x}^{\text{patch}}_{r, c}$, with $r$ and $c$ representing the row and column coordinates, respectively. The dispatcher sequences two patches for the prefetching and inference branches. Within the DN layer, the prefetching branch calculates and caches statistical moments. For the inference branch, statistics for the patch and its eight surrounding patches are queried. Subsequently, fast interpolations are employed to estimate the mean ($\hat{\boldsymbol{\mu}}^{\text{pixel}}_{c,r}$) and standard deviation ($\hat{\boldsymbol{\sigma^*}}^{\text{pixel}}_{c,r}$) for each pixel, facilitating dense normalization.}
\label{fig:framework}
\end{figure*}
}

\noindent\textbf{Ultra-high-resolution unpaired image-to-image translation.}
Performing unpaired image-to-image translation on ultra-high-resolution images is computationally expensive. The patch-wise-based method, which divides the input image into smaller patches and reassembles the translated ones, is a solution, but it often leads to tiling artifacts. To solve this problem, overlapping windows~\cite{lahiani2019virtualization, de2018stain} can be used, or a perceptual embedding consistency loss can be employed to learn color, contrast, and brightness invariant features~\cite{lahiani2020seamless}. Meanwhile, downsampling-based methods~\cite{song2022multi,liang2021high} avoid tiling artifacts but may result in detail loss and increased spatial complexity in upsampled images. Thumbnail Instance Normalization (TIN)~\cite{chen2022towards} eliminates gap-type tiling artifacts by assuming that all patches share the same global image-level statistics, but may result in over/under-colorizing and jitter-type tiling artifacts. Kernelized Instance Normalization (KIN)\cite{ho2022ultra} involves a two-stage pipeline and computes patch-level statistics using convolution operations to preserve local information but requires selecting an optimal kernel. Our DN differentiates itself by estimating pixel-level statistics to reduce tiling artifacts ($\mathcal{P}_1$) and preserve local hues and colors ($\mathcal{P}_2$) in a single pass ($\mathcal{P}_3$), without the need for hyperparameter tuning ($\mathcal{P}_4$).

{
\setlength\abovecaptionskip{-0.5ex}
\setlength\belowcaptionskip{-4.0ex}
\begin{figure}[t]
\begin{center}
\includegraphics[width=1.0\linewidth]{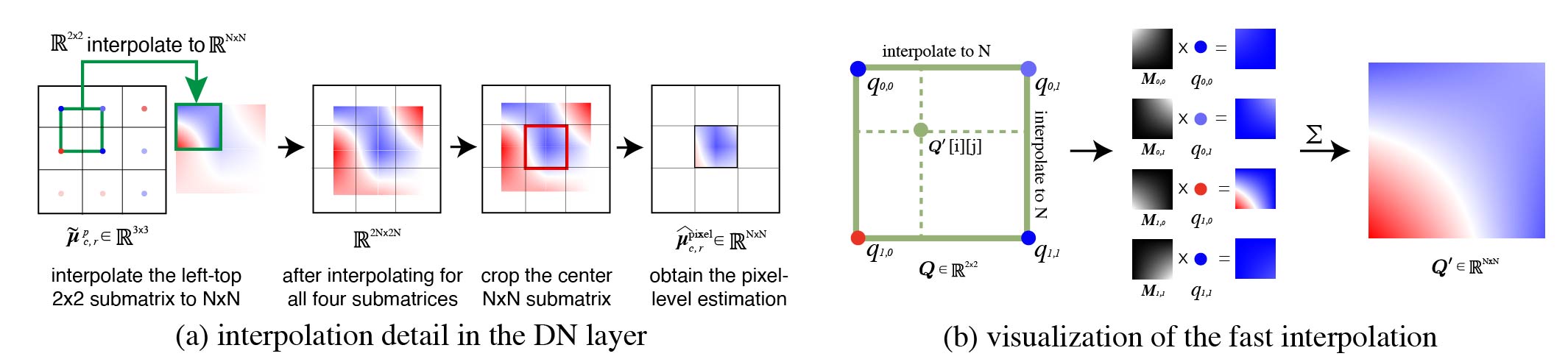}
\end{center}
   \caption{\textbf{Details of the fast interpolation operation utilized in DN.} Panel (a) illustrates the process of deriving N$\times$N pixel-level statistical moment estimations from a 3$\times$3 matrix. Panel (b) visualizes the matrix multiplication operation involved in fast interpolation.}
\label{fig:biin_detail}
\end{figure}
}

\section{Proposed methods}
\subsection{Overall framework}
Unpaired I2I translation aims to train a generator $\mathcal{G}$ to translate an image $\boldsymbol{X}$ in domain $\mathcal{X}$ to another domain $\mathcal{Y}$ even when there are no corresponding paired images. The output of $\mathcal{G}$ is denoted as $\hat{\boldsymbol{Y}}$ and is expected to be in domain $\mathcal{Y}$. In the context of UHR unpaired I2I translation, all images in domain $\mathcal{X}$ have a high resolution of $H\times W$. To enable $\mathcal{G}$ to handle images with infinite resolution, the model must be trained and executed in a patch-wise manner to reduce the GPU space complexity to a constant.

After training an I2I generator $\mathcal{G}$ with patches, the IN layers are replaced with DN layers. During the inference process, an UHR image $\boldsymbol{X} \in \mathbb{R}^{H \times W}$ is divided into patches $\boldsymbol{x}^{\text{patch}}_{c,r} \in \mathbb{R}^{N \times N}$, each with a size of $N \times N$, and their coordinates ($c$, $r$) relative to the original image are recorded. Here, $c \in \{0, 1, ..., \lceil \frac{H}{N} \rceil - 1\}$ and $r \in \{0, 1, ..., \lceil \frac{W}{N} \rceil - 1\}$. Then, a \textit{dispatcher} sequentially inputs these patches and coordinates into the generator.

Fig.~\ref{fig:framework} illustrates the overall pipeline of our framework. Specifically, during each dispatch, two patches along with their coordinates are sent: one to the prefetching branch and the other to the inference branch. Both patches simultaneously go through all layers except the DN layer. In the DN layer, the patch in the prefetching branch undergoes a standard instance normalization~\cite{ulyanov2016instance}, storing the resultant statistics in the cache table ($T_{\mu}, T_{\sigma}$) with the coordinates as keys. The patch in the inference branch uses its coordinates to query the cache table for its own and its eight neighbors' statistics, forming two $3 \times 3$ matrices of coarse-level (patch-level) statistical moments ($\tilde{\boldsymbol{\mu}}^{\text{patch}}_{c,r}$, $\tilde{\boldsymbol{\sigma}}^{\text{patch}}_{c,r} \in \mathbb{R}^{3 \times 3}$). \textit{Fast interpolation} is then applied to these matrices to estimate fine-level (pixel-level) statistical measures ($\hat{\boldsymbol{\mu}}^{\text{pixel}}_{c,r}$, $\hat{\boldsymbol{\sigma^*}}^{\text{pixel}}_{c,r} \in \mathbb{R}^{N \times N}$), which are subsequently used for dense normalization. The translated patches are then reassembled into an UHR image, with the DN operation effectively reducing tiling artifacts.

\subsection{Details}
\noindent\textbf{Dispatcher.} Given a collection of patches along with their coordinates, the dispatcher first arranges them vertically to create a list of images, denoted as $P$. Specifically, $P$ is formed as $(\boldsymbol{x}^{\text{patch}}_{0,0}, \boldsymbol{x}^{\text{patch}}_{1,0}, ..., \boldsymbol{x}^{\text{patch}}_{h-1,0}, \boldsymbol{x}^{\text{patch}}_{0,1}, ... , \boldsymbol{x}^{\text{patch}}_{h-1,w-1})$, where $h=\lceil \frac{H}{N} \rceil$ and $w=\lceil \frac{W}{N} \rceil$. Subsequently, the dispatcher sequentially dispatches the images. At step $t$, the dispatcher sends out two images along with their coordinates: $P[t]$ for the inference branch and $P[t+h+2]$ for the prefetching branch. This arrangement ensures that the eight neighboring patches of $P[t]$, as originally cropped from the UHR image, have already been processed by the prefetching branch, guaranteeing that the corresponding statistical moments are cached in $T_{\mu}$ and $T_{\sigma}$ and can be queried. The iteration starts from $t=-(h+2)$ and goes up to $h\cdot w -1$. For $t$ values outside the range of sequence $P$, an empty image $\phi$ is provided, and the branch assigned to process it performs no action.

{
\setlength\abovecaptionskip{-0.5ex}
\setlength\belowcaptionskip{-4.0ex}
\begin{figure}[t]
\begin{center}
\includegraphics[width=1.0\linewidth]{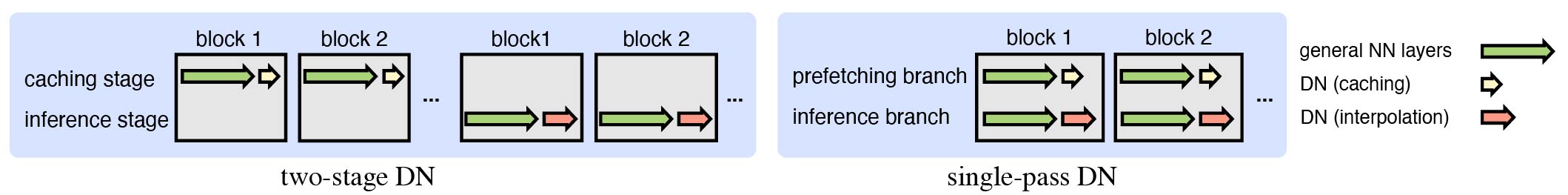}
\end{center}
   \caption{\textbf{Comparison of two-stage and single-pass DN.} A naïve implementation of DN might resemble KIN, operating in two stages. However, our dispatcher design and prefetching strategy enable the prefetching branch to run in parallel with the inference branch across most neural network (NN) layers, and to execute asynchronously in the DN layer, effectively hiding the runtime of the prefetching branch.}
\label{fig:parallelism}
\end{figure}
}

\noindent\textbf{Prefetching branch and caching.} When a patch $\boldsymbol{x}^{\text{patch}}_{c,r}$ enters the prefetching branch, it first undergoes a standard instance normalization process~\cite{ulyanov2016instance}.
{\small
\begin{equation}
    IN(\boldsymbol{x}^{\text{patch}}_{c,r}) = \gamma\left(\frac{\boldsymbol{x}^{\text{patch}}_{c,r}- \mathbb{E}[\boldsymbol{x}^{\text{patch}}_{c,r}]}{\sqrt {Var[\boldsymbol{x}^{\text{patch}}_{c,r}]}}\right) + \beta
\end{equation}
}

Here, $\mathbb{E}[\boldsymbol{x}^{\text{patch}}_{c,r}]$ and $\sqrt{Var[\boldsymbol{x}^{\text{patch}}_{c,r}]}$ are denoted as ${\mu}^{\text{patch}}_{c,r}$ and ${\sigma}^{\text{patch}}_{c,r}$, respectively. These represent the mean and standard deviation of $\boldsymbol{x}^{\text{patch}}_{c,r}$. Subsequently, these two statistics are stored in the cache table using their coordinates as keys; specifically, $T_{\mu}[c][r]:={\mu}^{\text{patch}}_{c,r}$ and $T_{\sigma}[c][r]:={\sigma}^{\text{patch}}_{c,r}$.

\noindent\textbf{Inference branch and dense normalization.} When a patch $\boldsymbol{x}^{\text{patch}}_{c,r}$ enters the inference branch, it first uses its coordinates to query the cache tables for its and its eight neighbors' statistical moments. Specifically, we query $T_{\mu}$ and $T_{\sigma}$ with keys $\{c-1, c, c+1\} \times \{r-1, r, r+1\}$, yielding two $3\times3$ matrices: $\tilde{\boldsymbol{\mu}}^{\text{patch}}_{c,r}$ and $\tilde{\boldsymbol{\sigma}}^{\text{patch}}_{c,r}$. Our goal is to derive two $N \times N$ pixel-level statistical moment estimations, $\hat{\boldsymbol{\mu}}^{\text{pixel}}_{c,r}$ and $\hat{\boldsymbol{\sigma^*}}^{\text{pixel}}_{c,r}$, from $\tilde{\boldsymbol{\mu}}^{\text{patch}}_{c,r}$ and $\tilde{\boldsymbol{\sigma}}^{\text{patch}}_{c,r}$, respectively, for $\boldsymbol{x}^{\text{patch}}_{c,r}$.

We assume that the patch-level statistics represent the statistics for the central pixel of the patch; for example, $\hat{\boldsymbol{\mu}}^{\text{pixel}}_{c,r}\left[\frac{N}{2}\right]\left[\frac{N}{2}\right] = {\mu}^{\text{patch}}_{c,r}$. Hence, we can utilize the process below to derive $\hat{\boldsymbol{\mu}}^{\text{pixel}}_{c,r}$ from $\tilde{\boldsymbol{\mu}}^{\text{patch}}_{c,r}$, with Fig.~\ref{fig:biin_detail}(a) providing a visual representation of the entire interpolation process.

\begin{enumerate}
    \item Perform fast interpolation on each corner of the $3\times3$ matrix $\tilde{\boldsymbol{\mu}}^{\text{patch}}_{c,r}$. For instance, for the top-left corner $\tilde{\boldsymbol{\mu}}^{\text{patch}}_{c,r}[0:1,0:1]$ (a $2\times2$ submatrix), apply fast interpolation.
    \item This interpolation is performed on a $2\times2$ submatrix to expand it to an $N\times N$ matrix.
    \item By interpolating each $2\times2$ submatrix into an $N\times N$ matrix, a larger $2N\times 2N$ matrix is constructed.
    \item The central $N\times N$ submatrix is then extracted from this $2N\times 2N$ matrix, serving as the pixel-level statistical estimation $\hat{\boldsymbol{\mu}}^{\text{pixel}}_{c,r}$ for the patch $\boldsymbol{x}^{\text{patch}}_{c,r}$. 
\end{enumerate}



For $\tilde{\boldsymbol{\sigma}}^{\text{patch}}_{c,r}$, we first calculate the inverse of each element to form $\tilde{\boldsymbol{\sigma}^*}^{\text{patch}}_{c,r}$. Then, the same interpolation and cropping processes are conducted to obtain the pixel-level statistical estimation $\hat{\boldsymbol{\sigma^*}}^{\text{pixel}}_{c,r} \in \mathbb{R}^{N\times N}$ for the patch $\boldsymbol{x}^{\text{patch}}_{c,r}$.

Now, we can utilize the pixel-level statistical moments to perform dense normalization, denoted as $DN(\cdot)$.

{\small
\begin{equation}
    DN(\boldsymbol{x}^{patch}, \hat{\boldsymbol{\mu}}^{\text{pixel}}_{c,r}, \hat{\boldsymbol{\sigma^*}}^{\text{pixel}}_{c,r}) = \gamma((\boldsymbol{x}^{patch} - \hat{\boldsymbol{\mu}}^{\text{pixel}}_{c,r}) \cdot \hat{\boldsymbol{\sigma^*}}^{\text{pixel}}_{c,r}) + \beta
\end{equation}
}

\noindent\textbf{Fast interpolation.} Different from interpolation in general cases where input and output sizes are always different, our DN requires computing interpolation from $\mathbb{R}^{2\times 2}$ to $\mathbb{R}^{N\times N}$ several times, with $N$ being a constant. Hence, we reformulate bilinear interpolation into fast interpolation, which can reduce computational demands and expedite DN computation.

Given a matrix $\boldsymbol{Q} \in \mathbb{R}^{2 \times 2}$, if we wish to interpolate it into $\boldsymbol{Q}' \in \mathbb{R}^{N \times N}$, using standard bilinear interpolation can be formulated as follows:
{\small
\begin{equation}
    \boldsymbol{Q}=
    \begin{bmatrix}
    q_{0,0} & q_{0,1} \\
    q_{1,0} & q_{1,1}
    \end{bmatrix}
\end{equation}
}

{\small
\begin{equation}
    \boldsymbol{Q}'[i][j] = \frac{1}{N^2}
    \begin{bmatrix}
    N - v_i & v_i 
    \end{bmatrix}
    \begin{bmatrix}
    q_{0,0} & q_{0,1} \\
    q_{1,0} & q_{1,1}
    \end{bmatrix}
    \begin{bmatrix}
    N - v_j \\
    v_j
    \end{bmatrix}, \quad \forall i,j \in \{0, 1, ..., N-1\}
\end{equation}
}
{\small
\begin{equation}
    \text{where}\ v_k = \frac{kN}{N-1}, \quad k \in \{0, 1, ..., N-1\}
\end{equation}
}

This can be further reformulated as:
{\small
\begin{align}
\boldsymbol{Q}'[i][j] &= \frac{1}{N^2}
\left\langle
\begin{bmatrix}
(N - v_i)(N-v_j) & (N-v_i) \cdot v_j \\
v_i \cdot (N-v_j) & v_i \cdot v_j
\end{bmatrix},
\begin{bmatrix}
q_{0,0} & q_{0,1} \\
q_{1,0} & q_{1,1}
\end{bmatrix}
\right\rangle \\
 &=
\left\langle
\begin{bmatrix}
\frac{(N - v_i)(N-v_j)}{N^2} & \frac{(N-v_i) \cdot v_j}{N^2} \\
\frac{v_i \cdot (N-v_j)}{N^2} & \frac{v_i \cdot v_j}{N^2} 
\end{bmatrix},
\begin{bmatrix}
q_{0,0} & q_{0,1} \\
q_{1,0} & q_{1,1}
\end{bmatrix}
\right\rangle, \forall i,j \in \{0, 1, ..., N-1\}
\end{align}
}



where $\left\langle,\right\rangle$ denotes the Frobenius inner product. It can be noted that for a given coordinate $(i, j)$, the interpolated value $\boldsymbol{Q}'[i][j]$ is a weighted sum of elements in $\boldsymbol{Q}$ with a fixed set of weights since $N$ is a constant.

Thus, the final interpolated result $\boldsymbol{Q}'$ can be written as:

{\small
\begin{equation}
    \boldsymbol{Q'} = \sum_{k=0}^{1}\sum_{l=0}^{1} q_{k, l} \cdot \boldsymbol{M}_{k, l}
\end{equation}
}

This equation represents the fast interpolation process (see Fig.~\ref{fig:biin_detail} (b)), where the elements in each matrix $\boldsymbol{M}_{k,l}\in \mathbb{R}^{N\times N}$ are defined as follows:

{\small
\begin{equation}
    \boldsymbol{M}_{0, 0}[i][j] = \frac{(N - v_i)(N-v_j)}{N^2}, \quad \forall i,j \in \{0, 1, ..., N-1\}
\end{equation}
}

{\small
\begin{equation}
    \boldsymbol{M}_{0, 1}[i][j] = \frac{(N-v_i) \cdot v_j}{N^2}, \quad \forall i,j \in \{0, 1, ..., N-1\}
\end{equation}
}

{\small
\begin{equation}
    \boldsymbol{M}_{1, 0}[i][j] = \frac{v_i \cdot (N-v_j)}{N^2}, \quad \forall i,j \in \{0, 1, ..., N-1\}
\end{equation}
}

{\small
\begin{equation}
    \boldsymbol{M}_{1, 1}[i][j] = \frac{v_i \cdot v_j}{N^2}, \quad \forall i,j \in \{0, 1, ..., N-1\}
\end{equation}
}

This reformulation highlights fast interpolation's desirable features (see Table~\ref{tab:ablation_algorithm}). First, it consists solely of matrix multiplication, which can be accelerated by a GPU. Second, all matrices $\boldsymbol{M}_{k, l}$ are consistent across all interpolation operations in our dense normalization, allowing for precomputation and caching.

\noindent\textbf{Parallelism and single pass.} Our dispatcher design obviates the need for separating caching and inference stages, enabling our framework to execute them concurrently in a single pass ($\mathcal{P}_3$). This efficiency is attributed to the inherent characteristics of GPUs. Specifically, processing a batch of images through a neural network layer (e.g., a convolutional layer) incurs a similar time cost regardless of the batch size. Consequently, two dispatched patches can be processed in parallel across most layers of the generator. Even though they perform different tasks upon reaching the DN layer, the operations are asynchronously enqueued and executed in parallel by the GPU. While data synchronization does require some time, it incurs only a minimal time cost. This parallel execution strategy allows the prefetching branch's operations to be effectively ``hidden'' beneath those of the inference branch, markedly reducing the overall runtime (see Fig.~\ref{fig:parallelism}).

\section{Experiments}
\subsection{Datasets}
\noindent\textbf{Natural images.}
To assess the effectiveness of DN, we utilized two publicly accessible datasets: Kyoto summer2autumn~\cite{ho2022ultra} and real2paint~\cite{yu2022towards}. Kyoto summer2autumn contains UHR unpaired images of summer and autumn landscapes (5,184$\times$3,456 pixels), useful for seasonal image conversion. The real2paint dataset contains UHR paintings by Vincent Van Gogh (4,000$\times$3,000 pixels to 10,000$\times$8,000 pixels) and real images (4,032$\times$3,024 pixels) from the UHDM dataset~\cite{yu2022towards}. Although low-resolution versions of Vincent Van Gogh paintings datasets are available~\cite{folego2016impressionism}, we collected 21 high-resolution images of public Vincent Van Gogh paintings online to facilitate research on UHR unpaired I2I translation. This curated list will be made publicly available.

\noindent\textbf{Pathological whole slide images (WSIs).}
In order to demonstrate the versatility of our DN module, we performed experiments on two additional pathological datasets for stain transformation. The ACROBAT dataset~\cite{weitz2022acrobat} consists of UHR WSIs of H\&E and corresponding estrogen receptor (ER), anti-progesterone receptor (PGR), human epidermal growth factor receptor 2 (HER2), and Ki67 WSIs. We randomly selected unpaired H\&E and PGR WSIs from this dataset as transformation targets. The ANHIR dataset~\cite{borovec2020anhir} contains WSIs from various organs with different staining and sizes ranging from 5,000$\times$5,000 pixels to 50,000$\times$50,000 pixels. For this dataset, we selected unpaired breast (H\&E to ER) and lung lesion (H\&E to Ki67) as our stain transformation targets.

\subsection{Experimental settings}
In our experiments with the aforementioned datasets, we cropped the UHR images into 512$\times$512 patches and trained CycleGAN, CUT, and L-LSeSim frameworks for 100 epochs with default hyperparameters. We replaced the IN layers with our DN layers for the inference process and compared the results with patch-wise IN, TIN, and KIN methods. The experiments were conducted on an NVIDIA RTX 3090 GPU. However, due to the GPU memory limitation, we were unable to directly use the IN layer with UHR images as input without cropping. We presented results obtained using the CUT model, with further findings available in the supplementary material.

\subsection{Metrics}
We compared the results from our DN, patch-wise IN, TIN, and KIN methods using qualitative and quantitative evaluation techniques. The translated UHR images were assessed using the standard Fréchet inception distance (FID)~\cite{heusel2017gans} metric. Additionally, we conducted a downstream task to differentiate between patches from the source and target domains. To explicitly showcase the adverse effects of tiling artifacts or over/under-colorization, we intentionally cropped new patches across the raw translated patches.%

However, since these metrics are not designed to evaluate UHR images and may not accurately identify tiling artifacts or over/under-colorizing, we conducted a detailed human evaluation consisting of three quality challenges. Participants were shown the source and translated images generated by DN, patch-wise IN, TIN, and KIN and asked to identify the image with the best quality, the fewest tiling artifacts, and the best color and hue. We recruited computer vision and pathology specialists in addition to the general population for the evaluation. For the translated WSIs, we conducted a fidelity challenge, following the AMT perceptual studies protocol~\cite{isola2017image}.

\section{Results}
{
\setlength\abovecaptionskip{-1.0ex}
\setlength\belowcaptionskip{-3.5ex}
\begin{figure}[t]
\begin{center}
\includegraphics[width=0.9\linewidth]{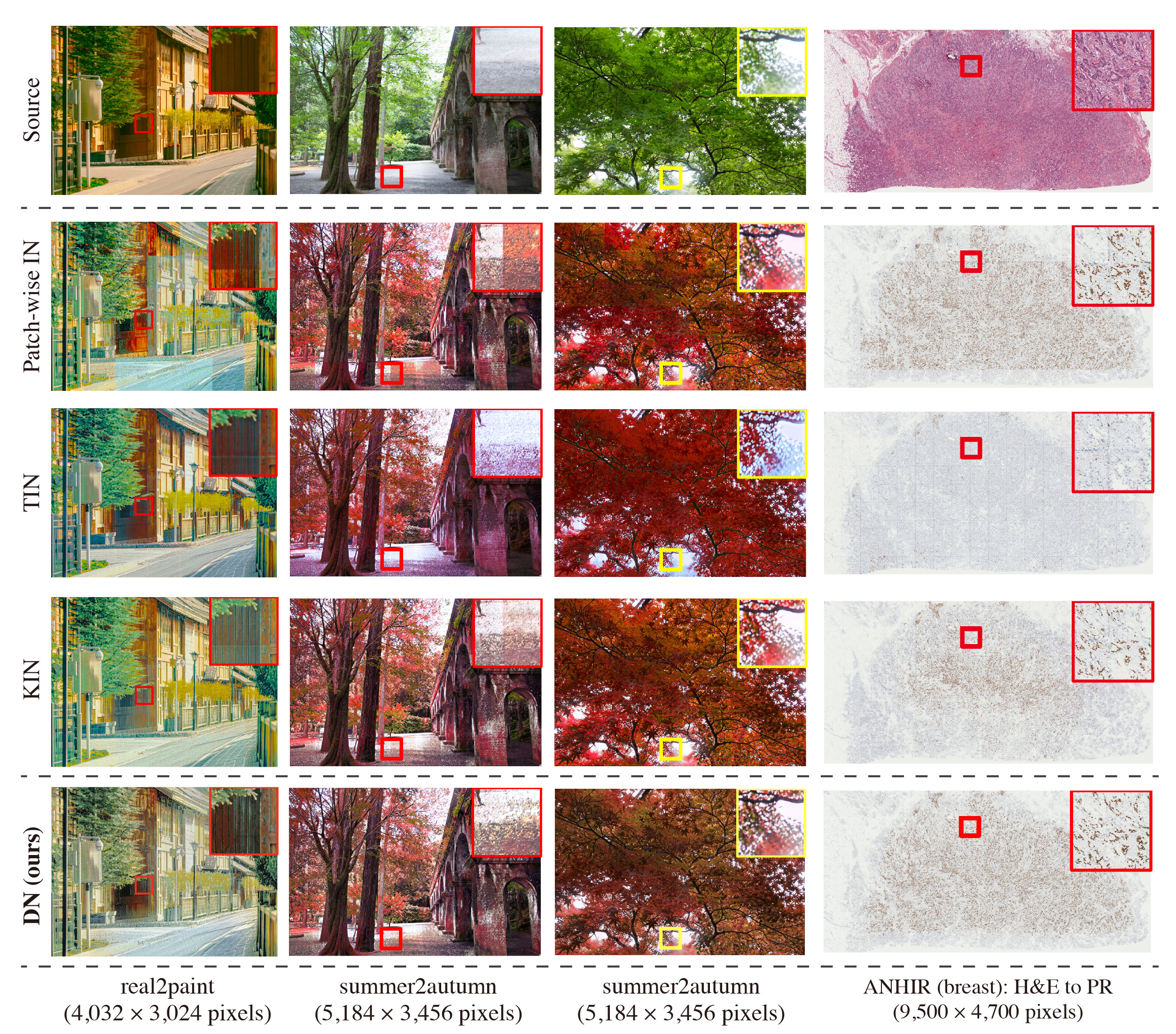}
\end{center}
   \caption{\textbf{Results of translated images.} The figure compares the translation results on UHR images using four normalization methods: patch-wise IN~\cite{ulyanov2016instance}, TIN~\cite{chen2022towards}, KIN~\cite{ho2022ultra}, and DN with a CUT~\cite{park2020contrastive} framework. Red close-up boxes highlight the comparisons of tiling artifacts, while yellow close-up boxes focus on evaluating over/under-colorizing and local hue preservation. DN shows the best performance overall. For a better view, please zoom in.} 
\label{fig:mix_compare}
\end{figure}
}

\subsection{Qualitative evaluation}
Fig.~\ref{fig:mix_compare}, Fig. S4, and Fig. S5 in the supplementary material show UHR images translated from natural and pathological WSI datasets. These images reveal that patch-wise IN generates a significant amount of gap-type tiling artifacts, while KIN mitigates some of these artifacts but at the expense of details, hue, and color. TIN reduces gap-type tiling artifacts but results in over/under-colorizing, loss of local hue details, and creates jitter-type tiling artifacts. Conversely, our DN approach is the only method that successfully diminishes tiling artifacts while maintaining local hue and color details, producing the best results.

\subsection{Quantitative evaluation}

\begin{table}[t]
\centering
\caption{\textbf{Quantitative Results.} The best-performing method in each experiment is highlighted in bold. DN surpasses both TIN~\cite{chen2022towards} and KIN~\cite{ho2022ultra}, as indicated by the underlined results. With respect to the FID metric, DN generally introduces the least disturbance to diminish tiling artifacts, in some cases even outperforming the intuitive lower bound. Across all experiments for the domain differentiation downstream task, measured by accuracy, DN consistently exceeds the performance of other methods, showcasing its superior efficacy in UHR image translation.}

\resizebox{0.9\columnwidth}{!}{%
\begin{threeparttable}
\label{tab:three_datasets}
\begin{tabular}{@{}lrrrrrrrr@{}}
\toprule
 &
  \multicolumn{4}{c}{\textbf{FID $\downarrow$}} & 
  \multicolumn{4}{c}{\textbf{Domain differentiation (\%) $\uparrow$}} \\
  \cmidrule(lr){2-5} \cmidrule(lr){6-9}
 &
\textbf{Lower bound\mbox{*}} &
\textbf{TIN} &
\textbf{KIN} &
\textbf{DN} &
\textbf{Patch-wise IN} &
\textbf{TIN} &
\textbf{KIN} &
\textbf{DN} \\
\midrule
\textbf{summer2autumn} & 98.281 & 117.268 & 98.003 & \textbf{\underline{97.732}} &
0.967 & 0.828 & 0.950 & \textbf{\underline{0.975}}\\
\textbf{real2paint} & 234.732 & 249.612 & 238.561 & \textbf{\underline{237.202}} &
0.971 & 0.556 & 0.757 & \textbf{\underline{0.986}}\\
\textbf{ACROBAT} & 21.046 & 43.988 & 27.224 & \textbf{\underline{21.346}} &
0.983 & 0.858 & 0.977 & \textbf{\underline{0.985}}\\
\textbf{ANHIR (breast)} & 64.202 & 161.128 & 91.443 & \textbf{\underline{68.616}} &
0.969 & 0.932 & 0.932 & \textbf{\underline{0.975}}\\
\textbf{ANHIR (lung lesion)} & 130.672 & 174.450 & 133.263 & \textbf{\underline{130.062}} &
0.880 & 0.863 & 0.880 & \textbf{\underline{0.900}}\\
\bottomrule
\end{tabular}%
\begin{tablenotes}
  \footnotesize
  \item Lower bound\mbox{*}: This is empirically achieved by patch-wise IN, as it is the optimized target of the backbone model, thereby setting an intuitive lower boundary for the FID values.
\end{tablenotes}
\end{threeparttable}
}
\label{tab:fid}
\end{table}

Table~\ref{tab:fid} (left part) presents the standard FID evaluation results on various datasets. Since the backbone model is optimized using patch-wise IN, this method can be considered as having the optimal FID values, thereby setting an intuitive lower bound, while other methods aim to minimally disturb the translation process to remove tiling artifacts. Overall, our hyperparameter-free DN ($\mathcal{P}_4$) outperforms TIN and KIN, indicating that it introduces the smallest adjustment necessary to achieve the goal, and local color and hue are preserved ($\mathcal{P}_1$). Remarkably, in some cases, it even surpasses the intuitive lower bound. On the other hand, KIN secures second place due to its balance between patch-wise IN and TIN. TIN inevitably yields the worst results because the use of global statistics introduces the largest disturbance.

Table~\ref{tab:fid} (right part) displays the results of the domain differentiation downstream task. DN consistently outperforms all other methods, likely indicating the involvement of the fewest tiling artifacts ($\mathcal{P}_2$) and the preservation of hue and color ($\mathcal{P}_1$). On the other hand, TIN yields the worst results, which is probably due to the large disturbances introduced by the use of global statistics.

To address the limitations of available metrics, we employed human evaluation to assess image quality (see Table~\ref{tab:human_general_and_profession_quality}). Three image quality challenges were conducted by forty participants, and our DN method achieved the best performance across all three challenges ($\mathcal{P}_1$ and $\mathcal{P}_2$), particularly for the Kyoto summer2autumn dataset. Additionally, we recruited eight computer vision specialists to evaluate the translation of natural images and eight pathology specialists to evaluate the results of stain transformation. Interestingly, the effectiveness of DN was more pronounced to these specialists. Furthermore, the fidelity challenge (see Fig. S8 in the supplementary material) revealed that the images generated by DN were nearly indistinguishable from real pathological images.

\begin{table}[h]
\centering
\caption{\textbf{Human evaluation results conducted by the general public and experts.} The best-performing method in each experiment is highlighted in bold. DN outperforms patch-wise IN~\cite{ulyanov2016instance}, TIN~\cite{chen2022towards}, and KIN~\cite{ho2022ultra}, as indicated by the underlined results. Overall, DN is the most preferred method across all aspects.}
\resizebox{\textwidth}{!}{%
\begin{threeparttable}
\begin{tabular}{@{}lcccccccccccccccccccccccc@{}}
\toprule
& 
  \multicolumn{8}{c}{\textbf{Has the best quality (\%) $\uparrow$}} & 
  \multicolumn{8}{c}{\textbf{Has the fewest tiling artifacts  (\%) $\uparrow$}} & 
  \multicolumn{8}{c}{\textbf{Exhibits the best color and hue  (\%) $\uparrow$}} \\ 
  \cmidrule(lr){2-9} \cmidrule(lr){10-17} \cmidrule(lr){18-25}
&
  \multicolumn{4}{c}{\textbf{By the general public}} & 
  \multicolumn{4}{c}{\textbf{By experts}} & 
  \multicolumn{4}{c}{\textbf{By the general public}} & 
  \multicolumn{4}{c}{\textbf{By experts}} & 
  \multicolumn{4}{c}{\textbf{By the general public}} & 
  \multicolumn{4}{c}{\textbf{By experts}} \\
  \cmidrule(lr){2-5} \cmidrule(lr){6-9} \cmidrule(lr){10-13} \cmidrule(lr){14-17} \cmidrule(lr){18-21} \cmidrule(lr){22-25}
&
  \textbf{IN\mbox{*}} &
  \textbf{TIN} &
  \textbf{KIN} &
  \textbf{DN} &
  \textbf{IN\mbox{*}} &
  \textbf{TIN} &
  \textbf{KIN} &
  \textbf{DN} &
  \textbf{IN\mbox{*}} &
  \textbf{TIN} &
  \textbf{KIN} &
  \textbf{DN} &
  \textbf{IN\mbox{*}} &
  \textbf{TIN} &
  \textbf{KIN} &
  \textbf{DN} &
  \textbf{IN\mbox{*}} &
  \textbf{TIN} &
  \textbf{KIN} &
  \textbf{DN} &
  \textbf{IN\mbox{*}} &
  \textbf{TIN} &
  \textbf{KIN} &
  \textbf{DN} \\
\midrule

\textbf{summer2autumn} & 10.00 & 15.79 & 12.11 & \underline{\textbf{62.11}} & 0.00 & 5.71 & 11.43 & \underline{\textbf{82.86}} & 4.74 & 18.95 & 12.63 & \underline{\textbf{63.68}} & 0.00 & 14.29 & 8.57 & \underline{\textbf{77.14}} & 6.32 & 15.26 & 12.63 & \underline{\textbf{65.79}} & 0.00 & 2.86 & 11.43 & \underline{\textbf{85.71}}\\

\textbf{real2paint} & 3.68 & 34.21 & 22.11 & \underline{\textbf{40.00}} & 0.00 & 28.57 & 22.86 & \underline{\textbf{48.57}} & 3.68 & 34.21 & 22.11 & \underline{\textbf{40.00}} & 0.00 & 22.86 & 22.86 & \underline{\textbf{54.29}} & 7.37 & 32.11 & 18.9 & \underline{\textbf{41.58}} & 0.00 & 25.71 & 20.00 & \underline{\textbf{54.29}}\\

\textbf{stain transformation} & - & - & - & - &
14.29 & 10.71 & 3.57 & \textbf{\underline{71.43}} &
- & - & - & - &
7.14 & 17.86 & 10.71 & \textbf{\underline{64.29}} &
- & - & - & - &
14.29 & 7.14 & 3.57 & \textbf{\underline{75.00}}\\
\bottomrule
\end{tabular}
\begin{tablenotes}
  \footnotesize
  \item IN\mbox{*}: patch-wise IN; -: not applicable
\end{tablenotes}
\end{threeparttable}
}
\label{tab:human_general_and_profession_quality}
\end{table}

\subsection{Runtime and resource utilization}
Table~\ref{tab:computation_comparison} presents the runtime and GPU VRAM usage for various methods. Employing operations on statistical moments generally leads to longer runtime but yields superior results. Distinctively, DN achieves faster execution in a single pass ($\mathcal{P}_3$) compared to KIN, with a modest increase in GPU VRAM usage. This efficiency is due to the parallel execution of prefetching and the inference branch, highlighting DN's innovative approach to parallelism design.

{
\setlength\abovecaptionskip{-1.0ex}
\setlength\belowcaptionskip{-3.5ex}
\begin{table}[h]
\centering
\caption{\textbf{Comparison of runtime and GPU memory usage.} Using an NVIDIA RTX 3090 GPU, we benchmarked the runtime and GPU VRAM usage for a 4,302 $\times$ 3,024 image. One-stage DN, despite involving substantial operations on statistical moments, runs faster than KIN.}
\resizebox{0.7\textwidth}{!}{%
\begin{threeparttable}
\begin{tabular}{@{}lrrrrr@{}}
\toprule
 &
\textbf{IN\mbox{*}} &
\textbf{TIN} &
\textbf{KIN} &
\textbf{DN} &
\textbf{DN} \\
\midrule
\textbf{Statistics type} & patch-level & image-level & patch-level & pixel-level & pixel-level \\
\textbf{\# of pipeline stage} & 1 & 1 & 2 & 2 & 1 \\
\textbf{Operations on statistics} &  &  & \Checkmark & \Checkmark & \Checkmark \\
\midrule
\textbf{Runtime (s)} & 2.46 & 2.62 & 4.42 & 5.51 & 4.35 \\
\textbf{GPU VRAM usage (mb)} & 2951 & 3335 & 3145 & 3161 & 4157 \\

\bottomrule
\end{tabular}%
\begin{tablenotes}
  \footnotesize
  \item IN\mbox{*}: patch-wise IN
\end{tablenotes}
\end{threeparttable}
}
\label{tab:computation_comparison}
\end{table}
}

\subsection{Ablation study}
{
\setlength\abovecaptionskip{-0.5ex}
\setlength\belowcaptionskip{-1ex}
\begin{figure}[t]
\begin{center}
\includegraphics[width=1.0\linewidth]{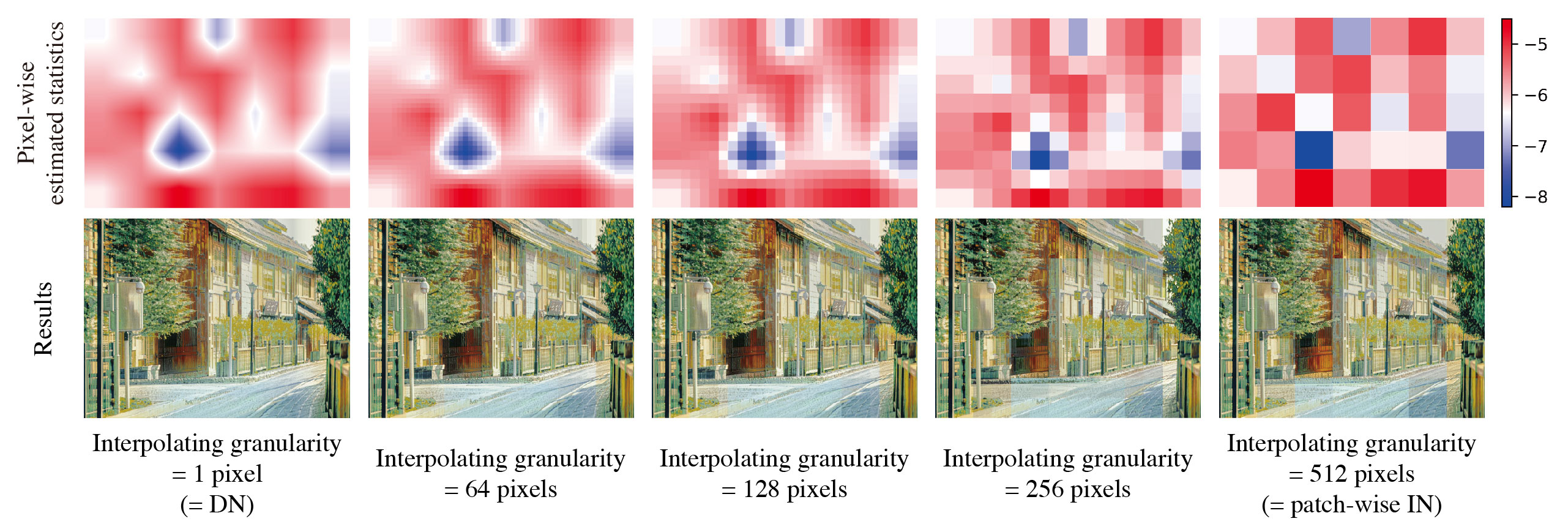}
\end{center}
   \caption{\textbf{Ablation study for interpolating granularity.} The gradual introduction of tiling artifacts is observed with increasing interpolating 
granularity. DN takes the most comprehensive approach to normalization by interpolating every pixel.}
\label{fig:ablation_biin}
\end{figure}
}
\noindent\textbf{Interpolating granularity.}
The statistical measures are interpolated by DN for each pixel, representing an interpolating granularity of 1 pixel. By incrementally increasing the interpolating granularity to 2, 4, and up to 512 pixels, gap-type tiling artifacts begin to emerge gradually, as illustrated in Fig.~\ref{fig:ablation_biin}. These results affirm that DN effectively diminishes tiling artifacts by the pixel-level statistical moment estimation.

\noindent\textbf{Runtime Optimization.} DN employs a fast interpolation algorithm and prefetching parallelism strategy to enable exhaustive estimation of pixel-level statistical moments efficiently. Table~\ref{tab:ablation_algorithm} presents evidence of significant acceleration. Performance benchmarks conducted on a single NVIDIA RTX 3090 GPU revealed that these strategies could achieve a speedup of 44 times for the entire image and 53 times per patch inference, respectively.

\begin{table}[h]
  \caption{\textbf{Speedup achieved using fast interpolation and prefetching parallelism.}  Benchmarking was conducted on an NVIDIA RTX 3090 GPU to evaluate runtime optimization strategies for a 4,302$\times$3,024 image. Although prefetching parallelism requires processing a slightly higher number of patches, it significantly enhances performance, achieving final speedups of 44 and 53 times for the entire image and per patch, respectively.}
  \label{tab:ablation_algorithm}
  \centering
  \resizebox{0.8\columnwidth}{!}{%
  \begin{tabular}{@{}ccc|cr@{}r@{}r@{}r@{}}
    \toprule
    \multicolumn{2}{c}{\textbf{Fast interpolation}} & \multicolumn{1}{c|} {\textbf{Prefetching}} & \multicolumn{1}{c}{\textbf{\# of}} &
    \multicolumn{1}{c}{\textbf{Runtime (s)}} & 
    \multicolumn{1}{c}{\textbf{Runtime (s)}} & 
    \multicolumn{1}{c}{\textbf{Speedup (times)}} &
    \multicolumn{1}{c}{\textbf{Speedup (times)}} \\

    \multicolumn{1}{c}{\textbf{Reformulation}} & \multicolumn{1}{c}{\textbf{Precomputation}} & 
    \multicolumn{1}{c|}{\textbf{parallelism}}&
    \multicolumn{1}{c}{\textbf{patches}}&
    \multicolumn{1}{c}{\textbf{(entire image)}}&
    \multicolumn{1}{c}{\textbf{(per patch)}}&
    \multicolumn{1}{c}{\textbf{(entire image)}}&
    \multicolumn{1}{c}{\textbf{(per patch)}}
    \\
    \midrule
    & & & 35 & 192.50 & 5.50 & 1x & 1x\\
    \Checkmark & & & 35 & 8.28 & 0.24 & 23x & 23x \\
    \Checkmark & \Checkmark & & 35 & 5.51 & 0.16 & 35x & 35x \\
    \Checkmark & \Checkmark &\Checkmark & 42 & 4.35 & 0.10 & 44x & 53x \\
    \bottomrule
  \end{tabular}
  }
\vspace{-1ex}
\end{table}

\section{Conclusion}
In this study, we have introduced DN for UHR unpaired I2I translation. DN estimates pixel-level statistical moments for normalization, thereby diminishing tiling artifacts and preserving local hue and color simultaneously. It can be seamlessly integrated into any unpaired I2I translation model equipped with IN layers, without necessitating model retraining or hyperparameter tuning. The proposed fast interpolation algorithm allows DN to efficiently estimate statistical moments for every pixel. Additionally, a prefetching parallelism strategy enables DN to operate in a single pass. Experimental results have demonstrated that DN outperforms all prior methods on datasets containing natural images and pathological WSIs. Furthermore, DN's ability to successfully perform stain transformation highlights its practicality in the medical domain.

\noindent\textbf{Limitations and discussion.} Although our research has demonstrated the superiority of DN over previous methods for UHR unpaired I2I translation, DN still requires patch-wise processing. Consequently, it would struggle to maintain the continuity of translated objects across patches. On the other hand, while less pronounced than TIN, jitter-type tiling artifacts occasionally emerge in the results, causing slight visibility issues and a lack of seamlessness. Addressing these limitations remains a goal for future work.

Furthermore, there is a lack of appropriate metrics and datasets for evaluating existing methods in UHR image translation. While we conducted human evaluations to mitigate this limitation, we recognize the importance of creating new metrics and releasing large datasets. To this end, we have released a curated list of the real2paint dataset to encourage further research into UHR unpaired I2I translation.

%
%
\bibliographystyle{splncs04}
\bibliography{main}
\end{document}


\title{Every Pixel Has its Moments: Ultra-High-Resolution Unpaired Image-to-Image Translation via Dense Normalization (Supplementary Material)} 

\titlerunning{Dense Normalization}

\author{Ming-Yang Ho\inst{1}\and
Che-Ming Wu\inst{2} \and
Min-Sheng Wu\inst{3} \and Yufeng Jane Tseng \inst{1}}

\authorrunning{Ho. et al.}

\institute{
National Taiwan University \and
Amazon Web Services \and
aetherAI\\
\email{kaminyou@cmdm.csie.ntu.edu.tw, unowu@amazon.com, vincentwu@aetherai.com, yjtseng@csie.ntu.edu.tw}
}

\maketitle

\begin{figure*}
\begin{center}
\includegraphics[width=0.95\linewidth]{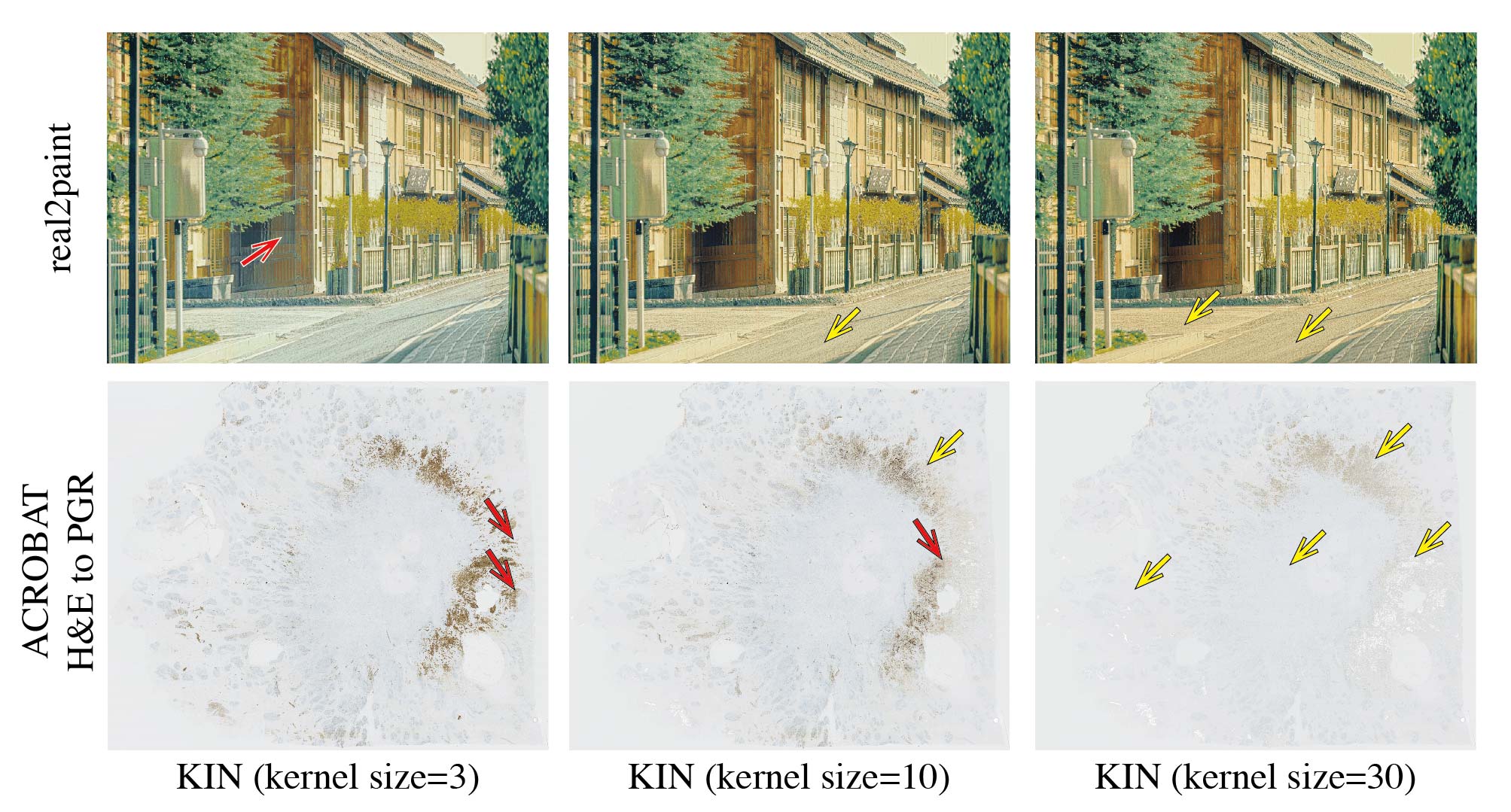}
\end{center}
   \caption{\textbf{The impact of varying kernel size in KIN [18].} Careful selection of kernel size is critical for balancing the removal of tiling artifacts and preservation of color and hue details to enhance the quality of translated images generated by KIN. Red arrows (\textcolor{red}{$\boldsymbol\searrow$}) indicate tiling artifacts, and yellow arrows (\textcolor{yellow}{$\boldsymbol\searrow$}) indicate over/under-colorizing.}
\label{fig:kin_kernel_size}
\end{figure*}

\begin{figure*}
\begin{center}
\includegraphics[width=0.95\linewidth]{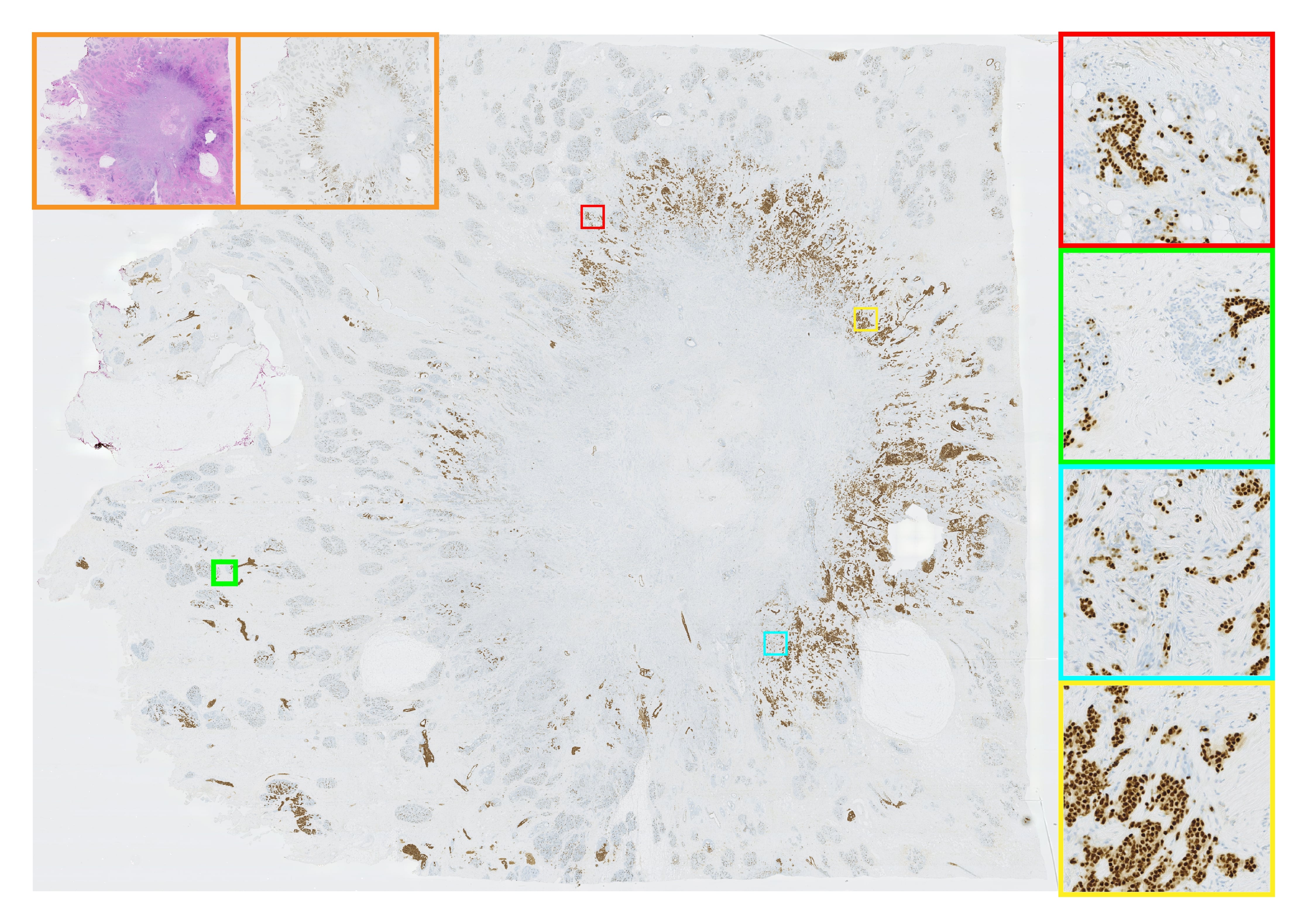}
\end{center}
   \caption{\textbf{A translated result of an ultra-high-resolution image (24,000$\times$28,000 pixels) generated using our Dense Normalization (DN).} Our DN is able to transform a whole slide image stained with hematoxylin and eosin (H\&E) into a progesterone receptor (PGR) stain without any discernible tiling artifacts, while also retaining all color and hue details. The right side of the image contains four close-up boxes for closer examination.}
\label{fig:acrobat_large}
\end{figure*}

\begin{figure*}
\begin{center}
\includegraphics[width=0.7\linewidth]{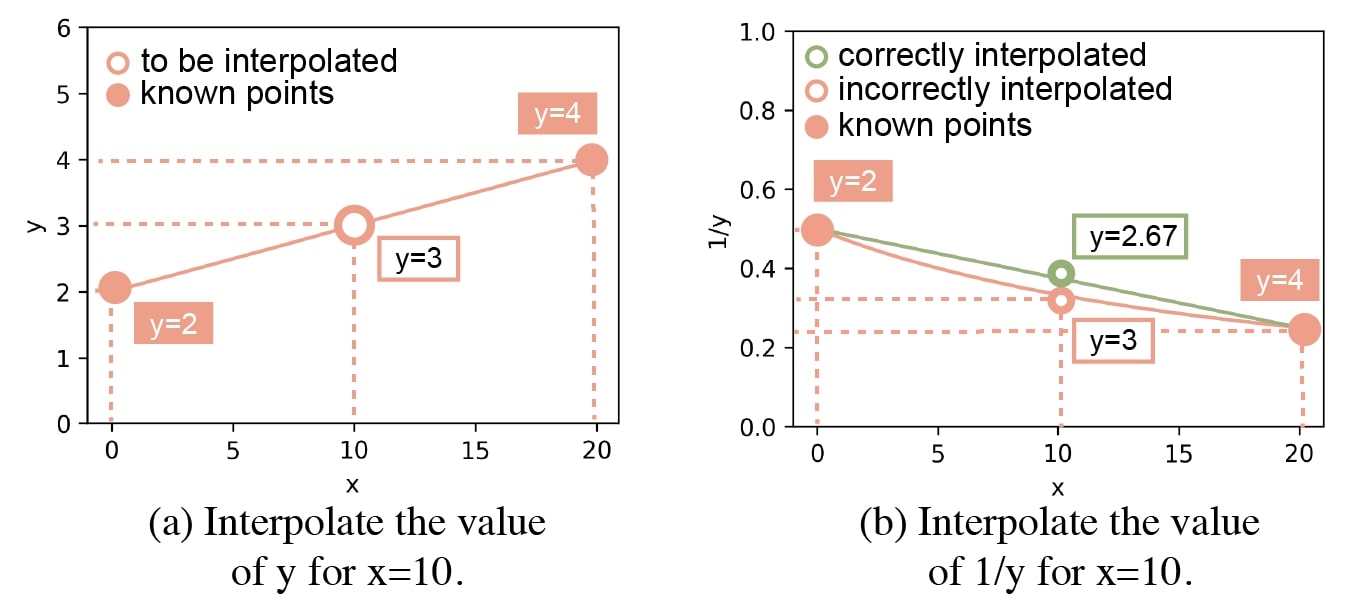}
\end{center}
   \caption{\textbf{Reciprocal-based interpolation.} These figures illustrate the challenges involved in computing interpolation on the reciprocal. In (a), two points (0, 2) and (20, 4) are given in the Cartesian coordinate system, and the interpolated value of $y$ at $x=10$ is obtained by taking the mean of 2 and 4, resulting in 3. In (b), when the interpolated results are transformed into the reciprocal form, they exhibit a hyperbolic function (\textcolor{orange}{orange line}). However, it is preferable for the reciprocal of the interpolated standard deviation to be a linear function (\textcolor{teal}{green line}) to prevent any nonlinear transformations from occurring during normalization of the image.}
\label{fig:reciprocal}
\end{figure*}

\begin{figure*}
\begin{center}
\includegraphics[width=0.94\linewidth]{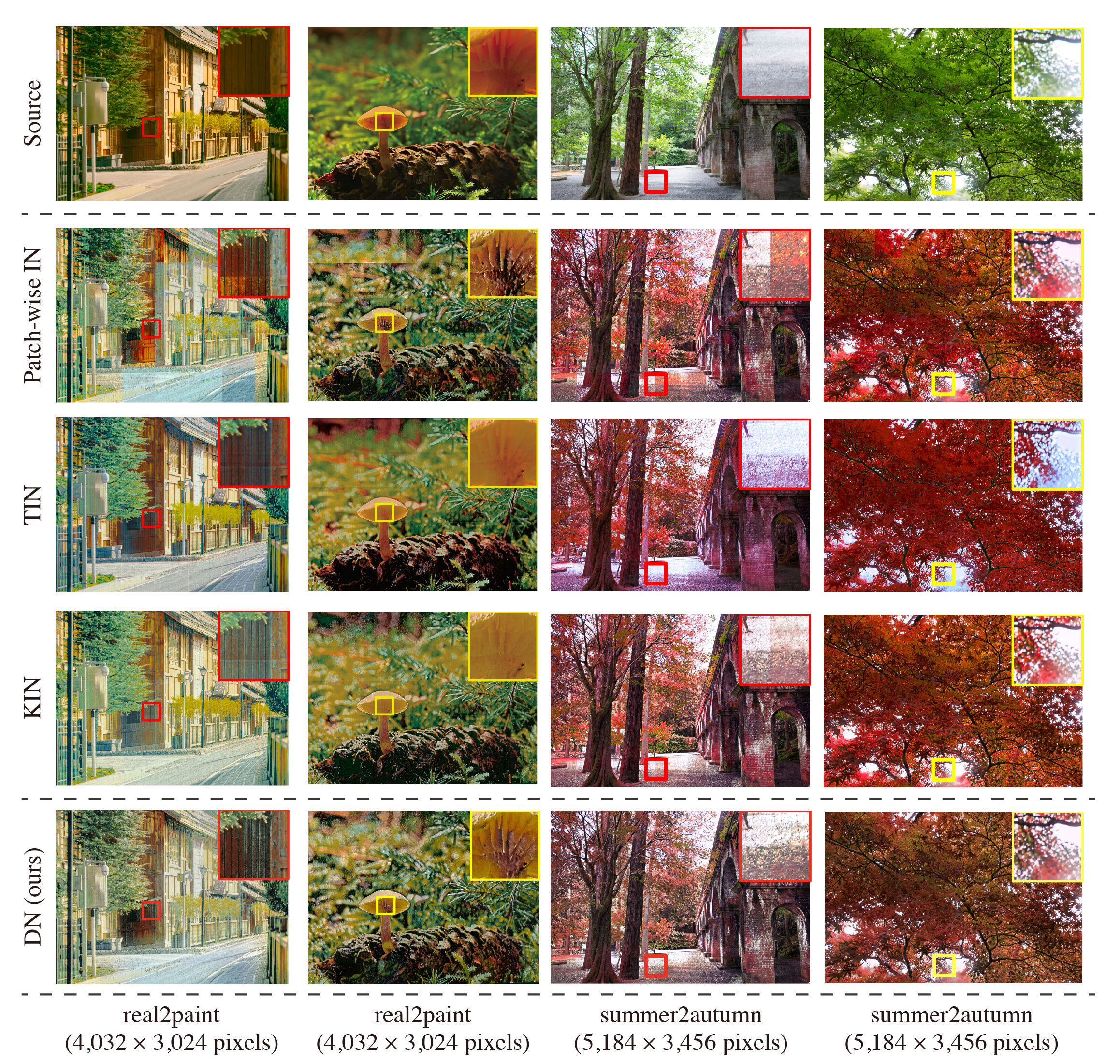}
\end{center}
   \caption{\textbf{Results of translation on natural images.} The figure compares the transformation results on UHR images using four normalization methods: patch-wise IN [17], TIN [19], KIN [18], and DN with a CUT [4] framework. Red close-up boxes highlight both tiling artifact comparisons, while yellow close-up boxes focus on evaluating over/under-colorizing and local hue preservation. DN shows the best performance overall. For a better view, please zoom in.} 
\label{fig:real_compare}
\end{figure*}

\begin{figure*}
\begin{center}
\includegraphics[width=0.94\linewidth]{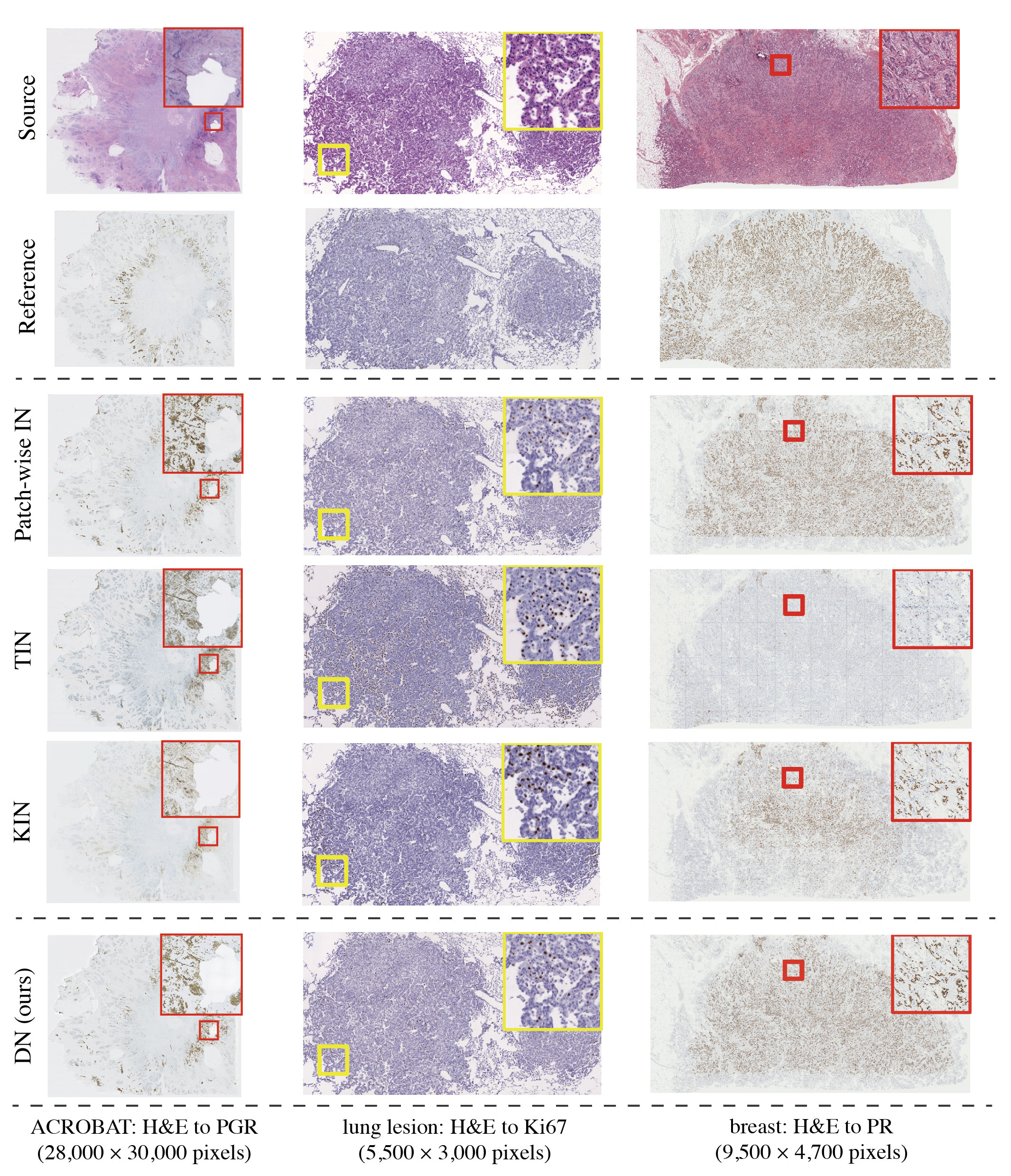}
\end{center}
   \caption{\textbf{Results of translation on pathological whole slide images.} The figure compares the stain transformation results on UHR whole slide images using four normalization methods: patch-wise IN [17], TIN [19], KIN [18], and DN with a CUT [4] framework. Red close-up boxes highlight both tiling artifact comparisons, while yellow close-up boxes focus on evaluating over/under-colorizing and local hue preservation. DN shows the best performance overall. For a better view, please zoom in.} 
\label{fig:patho_compare}
\end{figure*}

\begin{figure*}
\begin{center}
\includegraphics[width=0.95\linewidth]{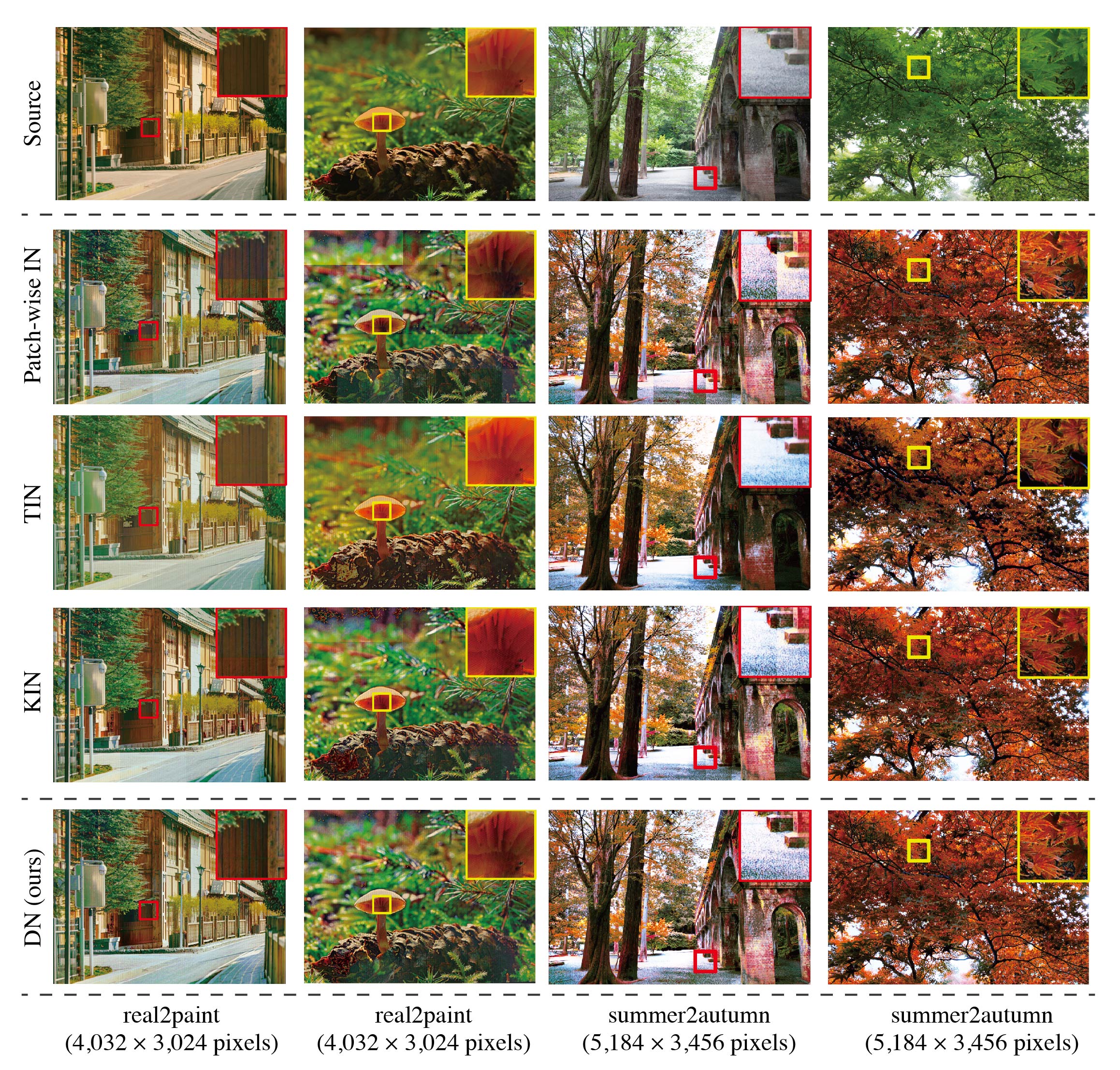}
\end{center}
   \caption{\textbf{Results of translation on natural images with a CycleGAN framework.} The figure compares the translation results on ultra-high-resolution images using four normalization methods: patch-wise IN [17], TIN [19], KIN [18], and DN with a CycleGAN framework. Red close-up boxes highlight both tiling artifact comparisons, while yellow close-up boxes focus on evaluating over/under-colorizing and local hue preservation. DN shows the best performance overall. For a better view, please zoom in.} 
\label{fig:real_compare_cyclegan}
\end{figure*}

\begin{figure*}
\begin{center}
\includegraphics[width=0.95\linewidth]{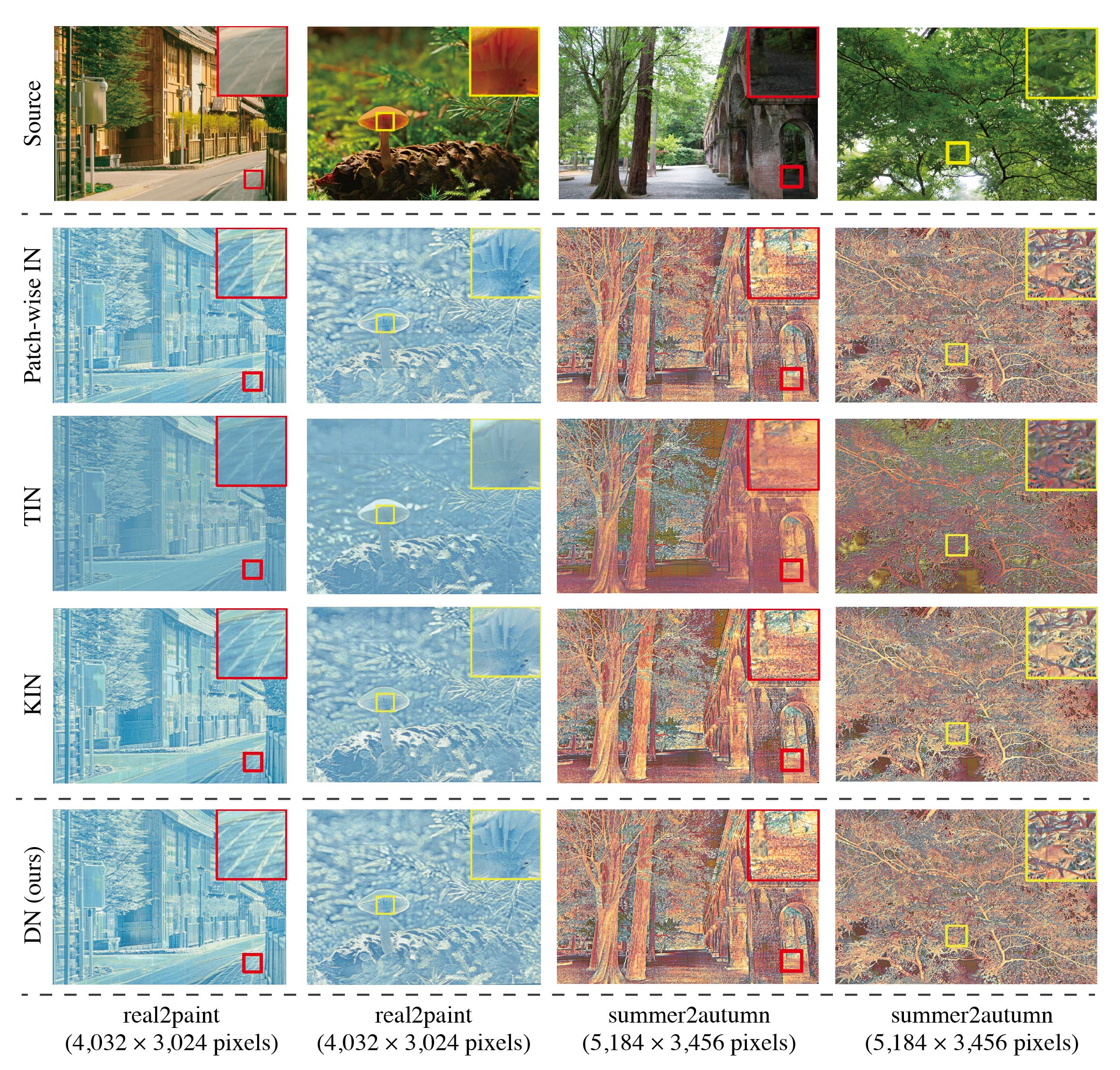}
\end{center}
   \caption{\textbf{Results of translation on natural images with an L-LSeSim framework.} The figure compares the translation results on ultra-high-resolution images using four normalization methods: patch-wise IN [17], TIN [19], KIN [18], and DN with an L-LSeSim framework. Despite the limitations of the L-LSeSim framework in effective translation, DN is still capable of removing tiling artifacts and maintaining color details. Red close-up boxes highlight both tiling artifact comparisons, while yellow close-up boxes focus on evaluating over/under-colorizing and local hue preservation. DN shows the best performance overall. For a better view, please zoom in.} 
\label{fig:real_compare_lsesim}
\end{figure*}

\begin{figure}[t]
\begin{center}
\includegraphics[width=0.4\linewidth]{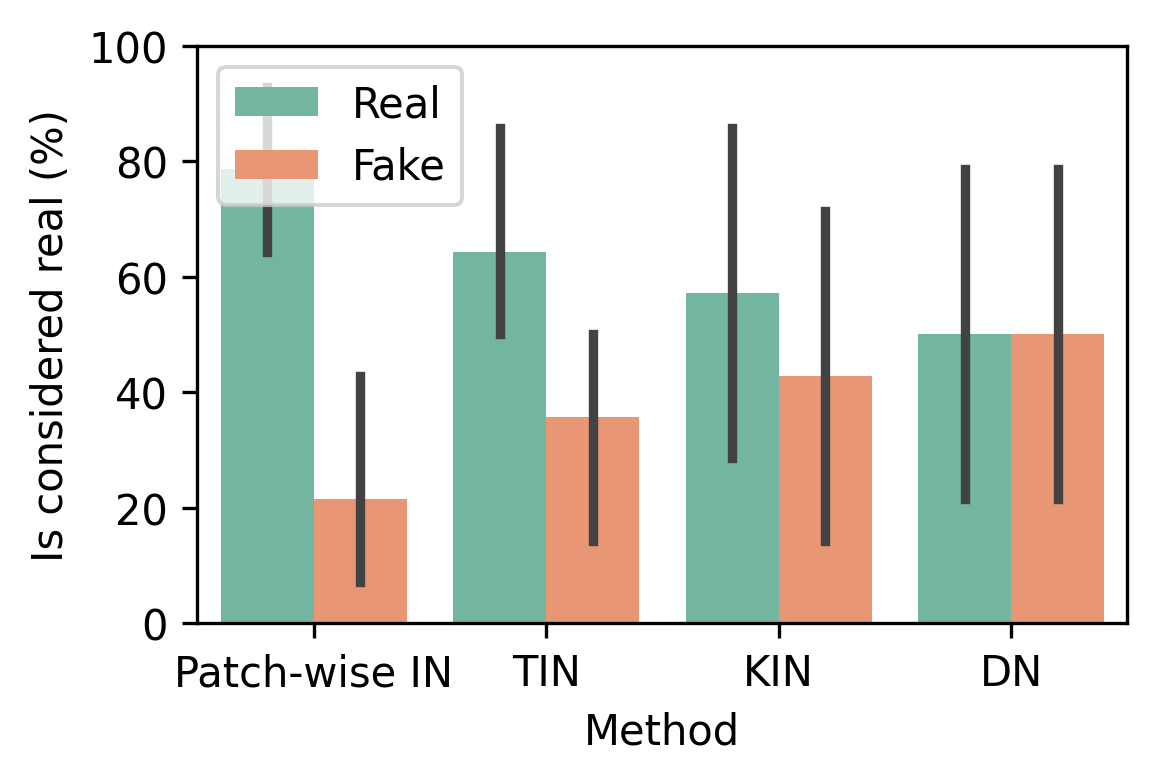}
\end{center}
   \caption{\textbf{Fidelity evaluation.} Images generated by DN are nearly indistinguishable from real pathological images.}
\label{fig:human_profession_fidelity}
\end{figure}

\subsection{Implementation details}
We implemented our Dense Normalization (DN) layer using PyTorch 1.13.0 and Python 3.9.5. All experiments were conducted on an Ubuntu 20.04.5 LTS operating system, equipped with an NVIDIA RTX 3090 GPU. For all experiments involving KIN, we used a constant kernel size of 5, as suggested by the authors of KIN. To reduce distortion in the margins during translation, each patch is initially reflectively padded, and then unpadded post-translation.

%
%
\clearpage
\bibliographystyle{splncs04}
\bibliography{main}